\documentclass[10pt,twocolumn,letterpaper]{article}

\usepackage{wacv}
\usepackage{times}
\usepackage{graphicx}
\usepackage{amsmath}
\usepackage{amssymb}
\usepackage{booktabs}
\usepackage{xcolor}
\usepackage{enumitem}
\usepackage{lipsum}
\usepackage[ruled,vlined,linesnumbered]{algorithm2e}
\usepackage{physics}
\usepackage{mathtools}
\usepackage{appendix}
\usepackage{float}
\usepackage{abstract}

\definecolor{codegreen}{rgb}{0.0,0.6,0.0}

\SetCommentSty{mycommfont}
\usepackage{float} 
\usepackage{caption}
\usepackage{datetime}


\usepackage{tabulary}
\usepackage{colortbl}
\usepackage[
    backend=bibtex,
    style=ieee,
  ]{biblatex}
\newcommand{\ob}{\cellcolor{orange!12}}
\newcommand{\rb}{\cellcolor{red!12}}
\newcommand{\bb}{\cellcolor{black!12}}
\newcommand{\dc}[1]{\textbf{\textsc{#1}}}

\newcommand*{\rowstyle}[1]{
  \gdef\xrowstyle{#1}%
  \xrowstyle\ignorespaces%
}

\newcolumntype{=}{
  >{\gdef\xrowstyle{}}%
}

\newcolumntype{+}{
  >{\xrowstyle}%
}

\usepackage[breaklinks=true,colorlinks,bookmarks=false]{hyperref}
\usepackage{caption}

\wacvfinalcopy 

\def\wacvPaperID{****} 

\renewcommand{\thefootnote}{\fnsymbol{footnote}}

\title{Strong-TransCenter: Improved Multi-Object Tracking based on Transformers with Dense Representations}

\begin{document}

\author{
Amit Galor\thanks{Corresponding author.} \hspace{1cm} Roy Orfaig \hspace{1cm} Ben-Zion Bobrovsky \vspace*{1mm}\\
School of Electrical Engineering, Tel-Aviv University \\
{\tt\small amitgalor@mai1.tau.ac.il \{royorfaig,bobrov\}@tauex.tau.ac.il}
}
\date{\vspace{-5ex}} 

\twocolumn[{%
\renewcommand\twocolumn[1][]{#1}%
\maketitle
}]
\renewcommand{\thefootnote}{\arabic{footnote}}
\begin{abstract}
Transformer networks have been a focus of research in many fields in recent years, being able to surpass the state-of-the-art performance in different computer vision tasks.  However, in the task of Multiple Object Tracking (MOT), leveraging the power of Transformers remains relatively unexplored. Among the pioneering efforts in this domain, TransCenter, a Transformer-based MOT architecture with dense object queries, demonstrated exceptional tracking capabilities while maintaining reasonable runtime. Nonetheless, one critical aspect in MOT, track displacement estimation, presents room for enhancement to further reduce association errors. In response to this challenge, our paper introduces a novel improvement to TransCenter. We propose a post-processing mechanism grounded in the Track-by-Detection paradigm, aiming to refine the track displacement estimation. Our approach involves the integration of a carefully designed Kalman filter, which incorporates Transformer outputs into measurement error estimation, and the use of an embedding network for target re-identification. This combined strategy yields substantial improvement in the accuracy and robustness of the tracking process. We validate our contributions through comprehensive experiments on the MOTChallenge datasets MOT17 and MOT20, where our proposed approach outperforms other Transformer-based trackers. The code is publicly available at: \url{https://github.com/amitgalor18/STC_Tracker}
\end{abstract}
\saythanks
\section{Introduction} \label{introduction}%
The task of multiple object tracking in video has been the center of much focus in computer vision research in the past few years, with various real-world applications (e.g. autonomous vehicles, sports analysis, surveillance, etc.). One of the most comprehensive benchmarks dedicated to this task is MOTChallenge~\cite{Milan2016}, having several datasets such as MOT17 and MOT20 which consist of fully annotated videos of pedestrians in various scenarios and environments. The goal of the tracker in the challenge is to infer the positions of each of the pedestrians in each frame while retaining their identities throughout the trajectories. There is a distinction in the challenge between online (causal) trackers which only use past and current frame information for the inference of a current frame and offline (non-causal) trackers that have access to information from all the video frames and can therefore infer on a current frame using future frame information.
Much research was done using these datasets \cite{Luo2021}, with many trackers using different approaches to complete the task. One of the major breakthroughs in the field was SORT (Simple Online and Realtime Tracking) \cite{Bewley2016}, having simplified the task and dividing it into several sub-tasks: Detection of all the pedestrians in the frame, association (Re-ID) between previous tracks (trajectories) and new detections, and prediction of track locations using a motion model.
The association sub-task was done using the well-known Hungarian Algorithm \cite{Kuhn1955}, and the track prediction was done using a standard Kalman filter \cite{Kalman_book}.
The method was further improved in the publication of DeepSORT \cite{Wojke2017}, which introduced the use of a neural network for the detection subtask.
Many more trackers in the following years used this Tracking-by-Detection paradigm \cite{Aharon2022,JDE_Wang2019,ByteTrack_Zhang2021,TMOH_2021}, which introduced mainly improvement to the detector network and the track management methods, while some trackers featured a separate network for occlusion handling \cite{CoupledNet2021}, advanced methods of assignment \cite{MAA_Stadler2022,UnsupTrack} or camera motion compensation \cite{Aharon2022,MAA_Stadler2022,Tracktor_Bergmann2019}.
Other trackers proposed to combine the detection with other components (e.g., appearance embedding, motion model, association) in one module \cite{JDE_Wang2019, Milan2017}.
In the last few years, several trackers \cite{Trackformer, TransTrack_Sun2020, MOTR_Zeng2021, Transcenter_Xu2022} introduced frameworks based on Transformer networks \cite{Attention} for MOT to benefit from the advantages of the generalizability of the model. The main advantage is the ability of the encoder-decoder architecture to encode features from the scene using a CNN (Convolutional Neural Network) \cite{ResNet2016} or PVT \cite{PVT_V2} while also encoding information about relationships between different parts of the scene and then decode the queries with a one-to-one assignment to objects. Transformer architectures have been shown to achieve better performance than CNN-based methods on several benchmarks, with great flexibility in the tasks they can complete \cite{DETR}. Nevertheless, with limited data, transformer-based trackers still fall short of surpassing trackers based on state-of-the-art object detectors, either from detection imprecision or inaccurate track motion prediction.

The main contributions of this paper are summarized as follows:
Our tracker is an attempt to improve upon a transformer-based tracker using post-processing methods for motion model estimation and re-identification. We show that a tuned Kalman filter performs better than the transformer motion tracking branch, and therefore reduces association errors. We introduce a modified implementation of the Kalman filter that is better optimized for the problem and includes integrated information from the network detection branch. As shown in \ref{results} it indeed achieves better results in the HOTA and IDF1 evaluation metrics and competitive results on the MOTA metric on both MOT17 and MOT20 test datasets.
\section{Related Work} \label{related_work}
\noindent \textbf{Transformers.}
In Multiple Object Tracking, the input is a sequence of frames. The encoder-decoder transformer architecture is therefore designed to encode the representations of the frames, use self-attention to reason about the objects in the scene and the encoder-decoder attention to access information from the whole frame.
The self attention helps preventing Identity Switches (IDSWs), terminate occluded tracks and initiate new tracks.
Both Transtrack \cite{TransTrack_Sun2020} and TrackFormer \cite{Trackformer} use sparse object queries to detect new objects and initialize tracking based on the DETR (Detection Transformer) \cite{DETR} architecture method, and track queries to keep information about the different objects across the frames, to achieve multi-frame attention. The most significant difference between the two is the association stage, in which TrackFormer took a point-based approach and TransTrack uses bounding-box-based association.
MOTR \cite{MOTR_Zeng2021} also employs a track query strategy but uses a temporal aggregation network to learn stronger temporal relations and obviate the need for IoU-based matching or Re-ID features.
TransCenterV2 \cite{Transcenter_Xu2022} took a different approach and achieved the best results out of all the Transformer-based trackers in the MOTChallenge benchmark \cite{Milan2016}, and therefore was chosen as the basis for our tracker. TransCenter trains pixel-wise dense queries to learn point-based tracking of pedestrian heatmap centers and sizes. They feed multi-scale tracking and detection queries into the decoder in order to find objects at different resolutions of the feature maps. It improved the efficiency by abandoning the heavy ResNet \cite{ResNet2016} based feature extraction and using the PVT (Pyramid Vision Transformer) \cite{PVT_V2} architecture as an encoder. TransCenter also uses pixel-level dense queries for detection, to avoid the insufficient number and the overlapping nature of sparse queries without positional correlations.
The track queries are kept active even when the object is not found for a few frames, in case it reappears after an occlusion. The main drawback is that the spatial information embedded into the query does not keep motion information from the past, limiting its accuracy and preventing the application for long term occlusions.
\noindent \textbf{Kalman Filter.}
Most of the trackers using the track-by-detection paradigm use the famous Kalman filter \cite{Kalman_book} to estimate the object's motion model and hence predict the location of the object in each frame. Kalman filter is an efficient, unbiased, optimal minimum-error estimator for linear dynamics.
The implementation of the Kalman filter that appeared in DeepSORT \cite{Wojke2017} was successfully used by many more modern trackers \cite{JDE_Wang2019, FairMOT, ByteTrack_Zhang2021, StrongSORT}, however this implementation can be improved with a careful choice of the state vector and measurement vector, as well as integration of additional information from the detector for the measurement noise estimation. Kalman filter allows a more precise association based on position, even in scenarios where the detector performs poorly, e.g., during occlusions or crowded scenes.
\begin{figure*}[h]
	\centering
	\includegraphics[width=1\textwidth,
    trim=0 100 0 500]{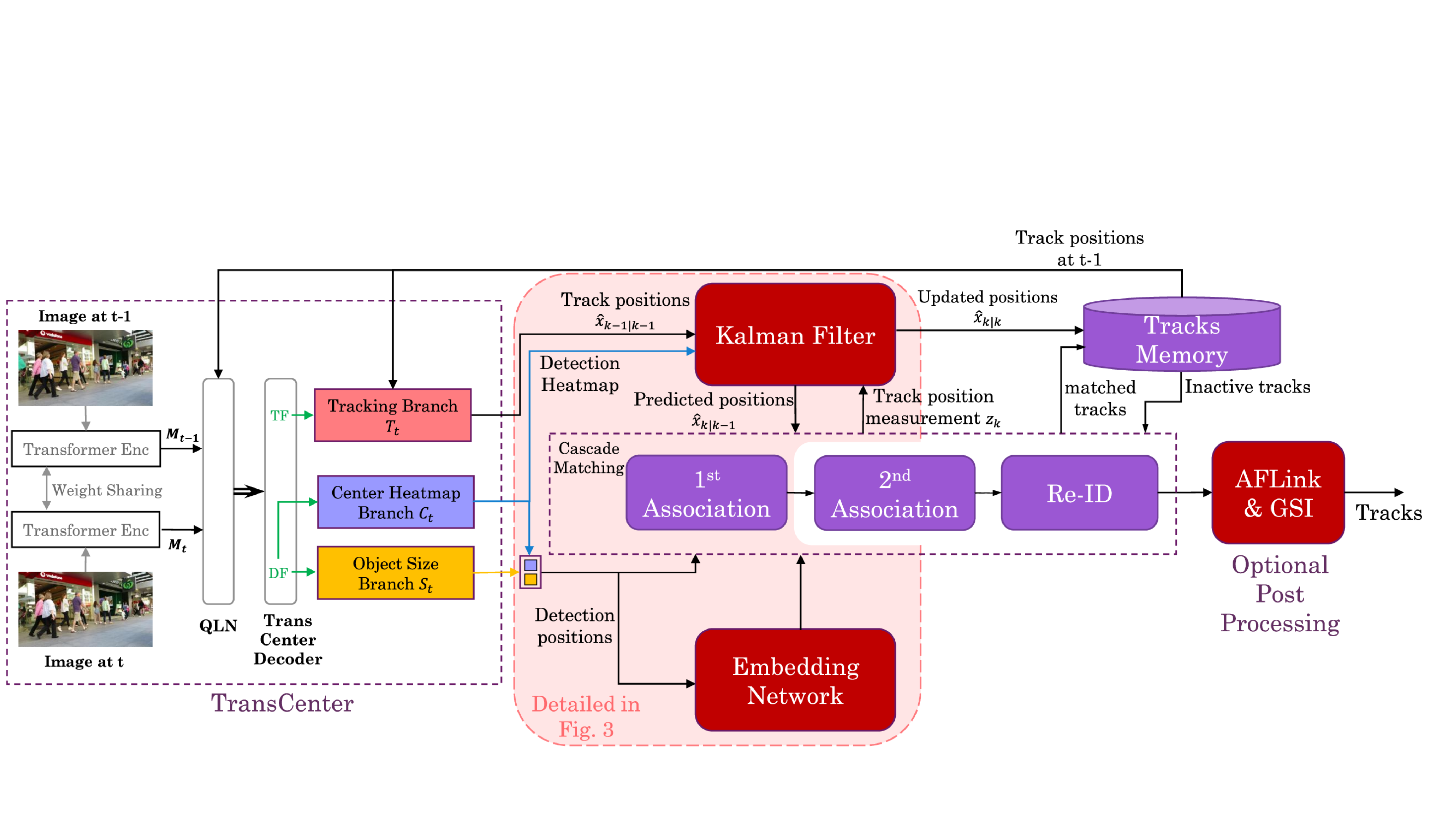}
	\caption{A flowchart overview of our STC tracker. The TransCenter \cite{Transcenter_Xu2022} main architecture was simplified on the left. The additional blocks are in dark red (Kalman filter \ref{kalman_filter} and Embedding Network \ref{embedding_network}) and the modified blocks are in purple. The detection and track positions are used to calculate the GIoU distances \cite{GIOU_Rezatofighi_2018_CVPR}, while the detection and track embeddings are used to calculate the appearance distance. The cascade matching contains two association steps that match new detections with existing tracks using a combined appearance and GIoU score. The Re-ID module attempts to match remaining detections with inactive tracks. The post-processing block is an optional addition, as in \cite{StrongSORT,ByteTrack_Zhang2021, Aharon2022} and described in \ref{offline}. The area in pink background is further detailed in Fig \ref{fig:association_flowchart}.} 
	\label{fig:flowchart}
	\vspace{1mm}
\end{figure*}
\noindent \textbf{Re-ID.}
Re-ID is the task of differentiating between different instances of the same class, and retaining the identity of an object seen in multiple occasions. Many works were conducted on optimal methods to represent an object's appearance, either by multiple views or with a single camera \cite{10032273,9106791,9229518,9911691,Person_reid_Ye2022}, by using graph collaboration, clustering methods or dedicated neural networks. Retaining an object's identity in a crowded scene is one of the main challenges in MOT. Several trackers use the same network for object detection and for extracting appearance features \cite{SOTMOT, JDE_Wang2019} in order to increase efficiency, while other methods use a separate deep neural network to extract features from the detected objects \cite{Wojke2017, StrongSORT}. Several methods train both networks together in order to achieve both efficiency and high performance \cite{RelationTrack, FairMOT,CSTrack, CorrTracker}. A different approach utilized a GAN (Generative Adversarial Network) to augment the expected pose of the pedestrians in each frame \cite{Kim2022}. The latest work \cite{Aharon2022} uses the Strong Baseline for person Re-identification \cite{SBS} and the FastReID pytorch library \cite{FastReID}, since this network achieves high performance on many person re-identification tasks. We chose this network since it was pre-trained on person re-identification datasets and has a simple implementation.

\section{Proposed Method} \label{proposed_method}
\subsection{Overview} \label{overview}
Our tracker, STC (Strong-TransCenter) is based on TransCenter \cite{Transcenter_Xu2022} with two main modifications integrated into the algorithm: A fine-tuned Kalman filter \cite{Kalman_book} and an Embedding Network based on FastReID \cite{FastReID}. A flowchart of our tracker pipeline is presented in Fig~\ref{fig:flowchart}.
The TransCenter main encoder-decoder architecture was kept as is. It takes as input two consecutive frames at a time and outputs the object location heatmaps with great performance. The modifications to the algorithm were focused on the track management, track localization and track-detection association. The cascade matching block contains two-step association depending on detection confidence as in \cite{ByteTrack_Zhang2021}, and a third Re-ID association. The first association stage matches active tracks in memory to new objects detected by the transformer with a high detection score. The second association stage matches the remaining active tracks with objects detected with low detection score. The third association matches the remaining detected objects with inactive tracks (tracks that were lost recently) and attempts to recover them back to the active tracks list. All three associations were adapted to include both a proximity score (GIoU - generalized Intersection over Union \cite{GIOU_Rezatofighi_2018_CVPR}) and an appearance score based on embedding from an Embedding network. An optional offline module was added and discussed in \ref{offline}.

\subsection{Kalman Filter} \label{kalman_filter}
\begin{figure}[htp]
        \centering
        \includegraphics[width=1\linewidth]{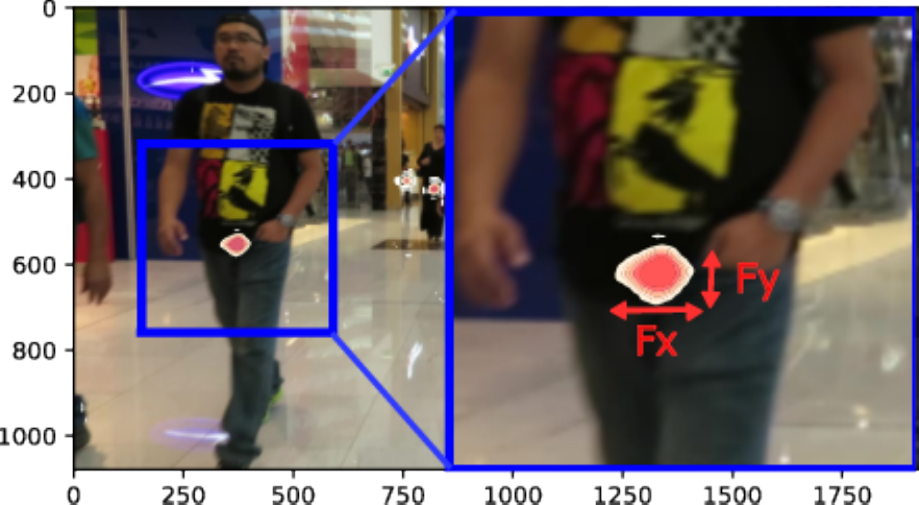}
        \caption{Visualization of the FWHM calculation. The TransCenter Transformer predicts a heatmap of the object detections. The blue window on the right is a zoom-in on an area in the left of the frame. The width of each peak (FWHM) is calculated in the x and y direction, and is denoted by Fx and Fy.}
        \label{fig:fwhm_vis}
\end{figure}

Kalman filter in a tracking task is used on trajectories to propagate each track's location to the next frame (Kalman predict phase), while the Kalman motion model is updated every time a track is associated to a new detection (Kalman update phase). Given an estimated state vector: 
\begin{equation}
\begin{aligned}
\mathbf{x}_k = \mathbf{\phi}_k\mathbf{x}_{k-1} +\mathbf{w}_k
\end{aligned}
\label{eq:KF_estimated_state}
\end{equation}
Where $\mathbf{\phi}_k$ is the state transition matrix and $\mathbf{w}_k$ is a white processing noise, and an observation:
\begin{equation}
\begin{aligned}
\mathbf{z}_k = \mathbf{H}_k\mathbf{x}_k +\mathbf{v}_k
\end{aligned}
\label{eq:KF_observation}
\end{equation}
Where $\mathbf{z}_k$ is a measurement at frame $k$, $\mathbf{H}_k$ is the observation matrix (the ideal relation between measurement and state vector), and $\mathbf{v}_k$ is a white measurement noise, the two steps can be represented by these recursive equations for each frame $k \in \mathbb{N}$:

\noindent \textbf{predict phase:}
\begin{equation}
\begin{aligned}
    & \hat{\mathbf{x}}^-_k = \mathbf{\phi}_k \hat{\mathbf{x}}_{k-1}\\
    & \mathbf{P}^-_k = \mathbf{\phi}_k \mathbf{P}_{k-1} \mathbf{\phi}_k^\top + \mathbf{Q}_k
\end{aligned}
\label{eq:predict_kf}
\end{equation}
\textbf{update phase:}
\begin{equation}
    \begin{aligned}
    & \mathbf{K}_k = \mathbf{P}^-_k \mathbf{H}_k^\top (\mathbf{H}_k \mathbf{P}^-_k \mathbf{H}_k^\top + \mathbf{R}_k)^{-1} \\
    & \hat{\mathbf{x}}_k = \hat{\mathbf{x}}^-_k + \mathbf{K}_k (\mathbf{z}_k - \mathbf{H}_k \hat{\mathbf{x}}^-_k) \\
    & \mathbf{P}_k = (\mathbf{I}- \mathbf{K}_k \mathbf{H}_k) \mathbf{P}^-_k
    \end{aligned}
    \label{eq:update_kf}
\end{equation} 
Where $\hat{\mathbf{x}}^-_k$ is the prior estimate, $\hat{\mathbf{x}}_k$ is the updated estimate, $\mathbf{P}_k$ is the updated estimate covariance matrix, $\mathbf{P}^-_k$ is the prior estimate covariance matrix $\mathbf{K}_k$ is the Kalman gain, $\mathbf{Q}_k$ is the process noise covariance, and $\mathbf{R}_k$ is the measurement noise covariance.
The implementation we chose was based on the work of DeepSORT \cite{Wojke2017} but the state vector and measurement representation were amended:

\begin{equation}
\label{eqn:kf_state}
\mathbf{x} =  [x_c, y_c, w, h, \dot{x_c}, \dot{y_c}, \dot{w}, \dot{h}]^\top
\end{equation}
Where ($x_c$, $y_c$) are the 2D coordinates of the object center in the image plane.  $w$ is the bounding box width and $h$ is the bounding box height. We replaced the aspect ratio in the state vector with the object width, to avoid non-linearity caused by division by small numbers.
Many works \cite{Wojke2017, FairMOT,StrongSORT, ByteTrack_Zhang2021} used an implementation that converts the measurement vector from the detector network, usually taking the form:
\begin{equation}
\label{eqn:zeta_vector}
\mathbf{\zeta} =  [x_{tl}, y_{tl}, x_{br}, y_{br}]
\end{equation}
to the same form of the state vector:
\begin{equation}
\label{eqn:zed_vector}
\mathbf{z} =  [x_c, y_c, a, h]
\end{equation}
Where ($x_{tl}$, $y_{tl}$) are the top left corner coordinates and ($x_{br}$, $y_{br}$) are the bottom right corner coordinates.
Accordingly, the common implementations choose the observation matrix to be the trivial projection matrix from state space to measurement space for the 4 position arguments, and zeros for the 4 velocity arguments:
\begin{equation}
\label{eqn:original_H}
\tilde{\mathbf{H}} =  [\mathbf{I}]_{4\times4} [\mathbf{0}]_{4\times4} 
\end{equation}
However, as seen in equation \ref{eq:KF_observation}, the measurement noise $\mathbf{v}_k$ is not included in this conversion. In order to get a more accurate estimation of the noise, we chose to use the original measurement such that $z=\zeta$ and the transformation will be done using the observation matrix:
\begin{equation}
\label{eqn:new_H}
\mathbf{H} =  \begin{bmatrix}
1 & 0 & -\frac{1}{2} & 0 & 0 & 0 & 0 & 0\\
0 & 1 & 0 & -\frac{1}{2} & 0 & 0 & 0 & 0\\
1 & 0 & \frac{1}{2} & 0 & 0 & 0 & 0 & 0\\
0 & 1 & 0 & \frac{1}{2} & 0 & 0 & 0 & 0
\end{bmatrix} 
\end{equation}
 Moreover, in order to get a more accurate estimate of the measurement error of the detector, we used the detection heatmap that is predicted by the transformer network before the creation of the detection bounding boxes. For each detection peak in the heatmap, we calculated the FWHM (Full Width Half Maximum) around the peak in the x and y directions, as demonstrated in Fig \ref{fig:fwhm_vis}. The heatmap can be correlated to a probability distribution, and it is known that for a Gaussian distribution with standard deviation $\sigma$:
 \begin{equation}
\label{eqn:fwhm}
\textnormal{FWHM} \propto \sigma
\end{equation}
 Therefore we modified $\mathbf{R}$, the covariance matrix of measurement noise $\mathbf{v}_k$, introduced in equation \ref{eq:KF_observation}, as follows:
\begin{equation}
\label{eqn:new_R}
\mathbf{R} =  \begin{bmatrix}
\alpha\cdot F_x^2 & 0 & 0 & 0\\
0 & \alpha\cdot F_y^2 & 0 & 0\\
0 & 0 & \alpha\cdot F_x^2 & 0\\
0 & 0 & 0 & \alpha\cdot F_y^2
\end{bmatrix} 
\end{equation}
Where $F_x$ and $F_y$ are the FWHM of the relevant detection in directions x and y respectively, and $\alpha=1$ is a constant, chosen after optimization. Different combinations of FWHM, constants, and the original width and height measures were tested and the best results were achieved with the simple approach presented in equation \ref{eqn:new_R}. We found that this modification is particularly helpful in scenes with a dynamic camera or objects with varying sizes such as those in the MOT17 dataset, and not very helpful in scenes with a static camera and objects of roughly the same size, such as those in the MOT20 dataset.
The integration of motion model estimation allows a more precise track position estimation and prevents misdetections and IDSWs, especially in cases where the detection is difficult.

\subsection{Embedding Network} \label{embedding_network}
In order to tackle the problem of IDSWs, especially in cases of crowded scenes with many possible matches in close proximity, we incorporated a method to base association on appearance instead of only GIoU. The implementation we chose was the FastReID library \cite{FastReID}, based on the SBS architecture \cite{SBS} with ResNeSt50 backbone \cite{ResNeSt}. The model was trained on the MOT17 and MOT20 train sets as in \cite{Aharon2022}. Attempts were made to use the features derived from the TransCenter transformer itself as the object embeddings but it achieved poor results compared to the SBS network, which was trained specifically for the Re-ID task. The network creates an embedding vector to represent the image patch containing the detected object using the heatmap centers' locations from the transformer. After every matching with a high enough detection score, the embedding vector is updated as the new appearance of the associated track. This is because a low score detection from the transformer network usually correlates with a problematic visual detection, e.g., a partially occluded object, that does not represent the object's regular appearance. The update is applied using an EMA (Exponential Moving Average), as in \cite{Aharon2022} and \cite{JDE_Wang2019}. The embedding vector for track $i$ at frame $k$ will be therefore updated as such:
\begin{equation}
    \begin{aligned}
    e_i^k = \alpha e_i^{k-1} + (1 - \alpha) f_i^k
    \end{aligned}
    \label{eq:ema}        
\end{equation}
Where $f_i^k$ is the appearance embedding of the current matched detection and $\alpha=0.9$ is a momentum term. 
\begin{figure*}[htpb]
	\centering
	\includegraphics[width=1\textwidth,
    trim=0 0 0 300]{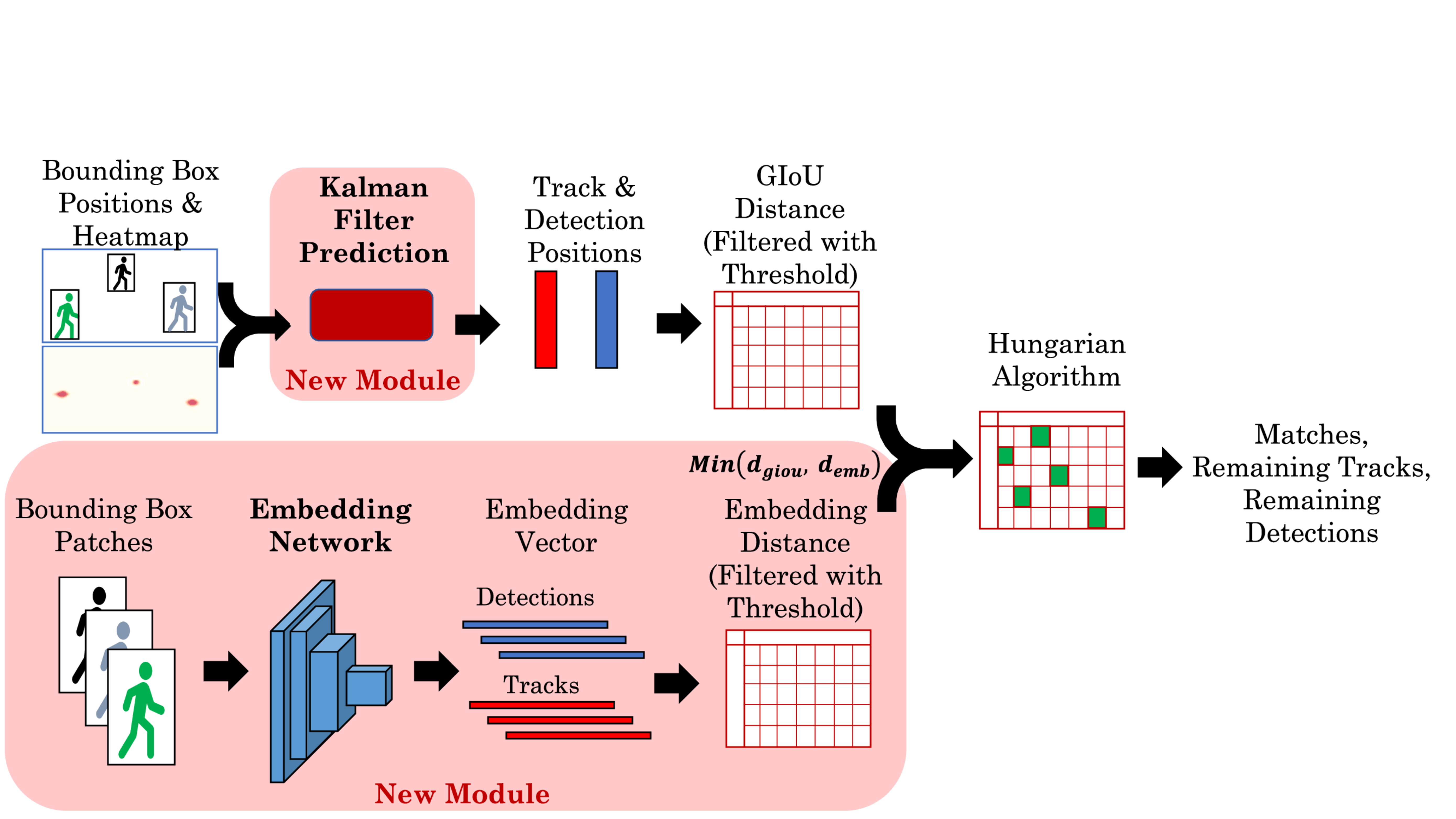}
	\caption{The main contributions of this paper: A flowchart of the modified association stage. The new modules are in the red background, and correspond to the Kalman, Embedding and 1st Association blocks from Fig \ref{fig:flowchart}. The fine-tuned Kalman filter uses information from the detection heatmap to improve the GIoU distance accuracy, while the appearance embedding distance helps in cases of pedestrians crowded together. The Hungarian algorithm for linear association is then applied to the resulting cost matrix. The process is then repeated for the 2nd Association and Re-ID stages, as seen in Fig \ref{fig:flowchart}.} 
	\label{fig:association_flowchart}
	\vspace{1mm}
\end{figure*}

As demonstrated in Fig \ref{fig:association_flowchart}, the difference between our suggested association and the association of the baseline Transcenter is both the integrated Kalman filter and the added Embedding network that allows a combined distance for matching. In every association stage an embedding distance is calculated between all the existing tracks and the new detections, as the cosine similarity between the track embedding $e_i^k$ and the new detection embedding vector $f_j^k$.
As in \cite{Aharon2022}, the combination of GIoU score and appearance embedding score was done by taking the minimum value between the two:
\begin{equation}
    C_{i, j} = min\{d_{i, j}^{giou}, \hat{d}_{i, j}^{emb}\}
    \label{eq:min_dist}        
\end{equation}
Where $C_{i, j}$ is the $(i, j)$ element of cost matrix $C$. $d_{i, j}^{giou}$ is the GIoU distance between the $i$-th track bounding box and the $j$-th detection bounding box, representing the motion cost. $d_{i, j}^{emb}$ is the cosine distance between the average track appearance vector $i$ and the new detection appearance embedding $j$. $\hat{d}_{i, j}^{emb}$ is our new appearance cost. Before combining the distances we filter out potential matches that don't comply with chosen thresholds for appearance distance and GIoU distance. The chosen set of thresholds differs in our case since we used GIoU \cite{GIOU_Rezatofighi_2018_CVPR}) instead of the regular IoU. The thresholds were chosen empirically using a grid search.
The association itself, after creating the cost matrix, is done using the Hungarian algorithm \cite{Kuhn1955}.

\section{Experiments} \label{experiments}
\subsection{Dataset and Evaluation Metrics}
The experiments were conducted on the MOT17 dataset \cite{Milan2016} which is composed of 7 videos for training and 7 videos for testing, and the MOT20 dataset \cite{MOT20_Dendorfer2020} which is composed of 4 videos for training and 4 videos for testing. The videos feature many pedestrians in various natural scenarios (streets, shopping centers etc.) with different lighting conditions, static or moving camera, different camera fps and resolution etc. The train set contains annotation files with bounding box locations for every object in every frame. The test set has only the videos available while performance evaluation is done using the MOTchallenge website with restrictive measures to avoid overfitting.
Methods in the MOTChallenge are evaluated using several main metrics.

\textbf{MOTA} (Multipe Object Tracking Accuracy) \cite{MOTA_Bernardin2008} has been used as the main evaluation metric for MOT for many years. It is calculated by the formula: 
\begin{equation} 
    MOTA = 1-\frac{FN+FP+IDSW}{GT}
\label{MOTA_formula}
\end{equation}
Where $FN$, $FP$, $IDSW$, and $GT$ are the numbers of false negatives (misses), false positives, IDSWs and ground truth labels respectively in all the frames of the sequence. This metric places a greater emphasis on detection-based errors and is therefore more useful when evaluating a tracker's detector.

\textbf{MOTP} (Multiple Object Tracking Precision) \cite{MOTA_Bernardin2008} is the measure of position precision, regardless of of the detection and identification skills of the tracker. It is calculated as:
\begin{equation}
    MOTP = \frac{D}{M}
\label{MOTP_formula}
\end{equation}
Where $D$ is the sum of distances between predicted positions and ground truth positions, and $M$ is the number of matches found for all frames.

\textbf{IDF1} \cite{IDF1} is a measurement of the tracker's ability to retain all of the objects' identifications throughout the sequence, and is based on the standard F1 score that balances between precision and recall in classification tasks. It is widely used in the MOTChallenge benchmark and is calculated by the formula:
\begin{equation}
    IDF1 = \frac{2IDTP}{2IDTP+IDFP+IDFN}
\label{IDF1_formula}
\end{equation}
Where $IDFP$, $IDFN$ and $IDTP$ are identification false positive matches, false negative matches and true positive matches.

\textbf{HOTA} (Higher Order Tracking Accuracy) \cite{HOTA} is a relatively new evaluation metric that aims to balance between the ability of a tracker to detect all objects and associate their individual identifications in one number. It is calculated by the formula:
\begin{equation}
    \begin{aligned}
    & HOTA = \sqrt{\frac{\Sigma_c{\mathcal{A}(c)}}{\abs{TP}+\abs{FN}+\abs{FP}}} \\
    & \mathcal{A}(c) = \frac{\abs{TPA(c)}}{\abs{TPA(c)}+\abs{FNA(c)}+\abs{FPA(c)}}
    \end{aligned}
\label{HOTA_formula}
\end{equation}
Where $TP$, $FN$, and $FP$ are the detection true positives, false negatives, and false positives respectively, and each $TP$ of interest $c$ has a weighted metric $\mathcal{A}(c)$ that weighs the association true positives, false negatives and false positives $TPA$, $FNA$ and $FPA$, as thoroughly explained and illustrated in \cite{HOTA}. This evaluation method produces more intuitive performance scores in many scenarios, so we consider it a better evaluation metric for the MOT task.

\textbf{MT} (Mostly Tracked) and \textbf{ML} (Mostly Lost) metrics are intended to get an intuitive sense of the extreme cases of the best and worst trajectories to complement the MOTA metric which averages everything, but they are ambivalent to IDSWs. Therefore, like MOTA, these metrics give greater emphasis to the detector performance.

\subsection{Implementation} \label{implementation}
Our proposed method was implemented in Python and is available on GitHub. The inference run for our submitted MOTChallenge results was done on a computer with an Intel Xeon Silver 4110 CPU, and a Geforce GTX 1180 Graphics card with 8GB VRAM. We conducted the execution of the original TransCenter method on the same hardware configuration to recreate their results and directly compare it to our proposed method. This comparison encompassed performance metrics as well as runtime. Notably, our STC tracker demonstrated only a slight decrease in FPS (7\%-17\%) compared to the TransCenter baseline, primarily attributed to the additional embedding network. Running our tracker without the embedding network yielded the exact same FPS as the baseline, highlighting that our performance improvements did not significantly impact runtime efficiency.
Please note that we did not include an FPS (runtime) column in our results table due to variations in hardware configurations across different methods in the literature. While we executed our proposed STC tracker and the original TransCenter method on the same hardware for a direct performance comparison, many published methods employed significantly more powerful hardware setups. Therefore, adding FPS values to the table could lead to misleading comparisons.
The results rely on a series of thresholds for different parts of the pipeline. The high detection threshold for the first association stage is $0.3$ in MOT17 and $0.5$ in MOT20, and the low detection threshold for the second association stage is $0.1$, based on the detection confidence score as in \cite{Transcenter_Xu2022}. The threshold for the embedding distance and GIoU distance are $0.4$ and $1.0$ respectively. Based on the combined distance, the matching threshold for the linear association itself is $0.9$ for the first association and the Re-ID recovery association, and $0.4$ for the second association, since the second association considers detections with lower confidence.

\subsection{Results} \label{results}
Table I and Table II present results on the MOT17 and MOT20 test datasets, respectively. We can see that STC outperforms the trackers that used the same pretraining conditions in terms of the HOTA and IDF1 metrics by a great margin, while the MOTA metric, MT and ML are roughly the same as the baseline tracker. Our MOTP is similar to TransCenter as expected, since the track positions after the association stage are taken as the detection positions. The FN and FP measures can be quite noisy since they are very sensitive to tracking thresholds, thus they are not reliable evaluators by themselves, but the balance between them can indicate different tendencies of trackers to prefer misdetection over false-detection or vice versa. The combination of the lower IDSW rate and the similar FN rate compared to the baseline tracker demonstrates both the performance of the accurate detection process by the original TransCenter transformer \cite{Transcenter_Xu2022}, and the ability of the Kalman filter and improved association to track more consistently. It is worth noting that STC outperforms all of the transformer-based methods in the HOTA and IDF1 metrics.
\begin{table*}[htbp]
\begin{center}
    \caption{Results on MOT17 test set on public and private detections. The methods chosen for comparison use the same dataset for pretraining or additional datasets, indicated by the background color. The best result among methods trained in the same conditions is in \textbf{bold}.} 
    \label{table:mot17}
    \resizebox{\linewidth}{!}{
            \begin{tabular}{ l | c c c c c c c c c c|| c c c c c c c c c c}
            \toprule
            & \multicolumn{10}{c||}{Public Detections} & \multicolumn{10}{c}{Private Detections}\\\midrule
              Method & Data &  MOTA $\uparrow$ & MOTP $\uparrow$ & IDF1 $\uparrow$ & HOTA$ \uparrow$ & MT $\uparrow$ & ML $\downarrow$ & FP $\downarrow$ & FN $\downarrow$ & IDSW $ \downarrow$ &  Data &  MOTA $\uparrow$ & MOTP $\uparrow$ & IDF1 $\uparrow$ & HOTA$ \uparrow$ & MT $\uparrow$ & ML $\downarrow$ & FP $\downarrow$ & FN $\downarrow$ & IDSW $\downarrow$ \\ [0.5ex] 
             \midrule
                GSDT \cite{GSDT} & \bb & \bb & \bb &  \bb & \bb & \bb & \bb & \bb & \bb & \bb &  \rb \dc{5d1} & \rb 66.2 & \rb 79.9 & \rb 68.7 & \rb 55.5 & \rb 40.8 & \rb 18.3 & \rb 43,368 & \rb 144,261 & \rb 3,318 
                  \\ %
                
                SOTMOT \cite{SOTMOT} & \rb \dc{5d1} & \rb {{62.8}} & \bb & \rb \underline{\textbf{ 67.4}} & \bb & \rb {{24.4}} & \rb {{33.0}} & \rb \underline{\textbf{6,556}} & \rb {{201,319}} & \rb \underline{\textbf{2,017}}  
                  & \rb \dc{5d1} & \rb 71.0 & \bb & \rb 71.9 & \bb &  \rb 42.7  & \rb 15.3 & \rb 39,537 & \rb 118,983 & \rb 5,184  \\
                 
                GSDT\_V2 \cite{GSDT} &\bb&\bb&\bb&\bb&\bb&\bb&\bb&\bb&\bb & \bb&\rb\dc{5d1}& \rb73.2 &  \bb &  \rb66.5 & \rb 55.2 & \rb41.7 & \rb17.5 & \rb 26,397 &	\rb120,666	 & \rb3,891 
                 \\ %

                CorrTracker \cite{CorrTracker}& \bb & \bb & \bb &  \bb & \bb & \bb & \bb & \bb & \bb  & \bb & \rb \dc{5d1} & \rb \underline{\textbf{76.5}} & \bb & \rb 73.6 & \rb 60.7 & \rb 47.6 & \rb {{12.7}} & \rb 29,808 & \rb{ 99,510} & \rb 3,369 
                 \\ %
                 
                FairMOT \cite{FairMOT} & \bb & \bb & \bb &  \bb & \bb & \bb & \bb & \bb & \bb  & \bb& \rb \dc{5d1+CH} & \rb 73.7 & \rb 81.3 & \rb 72.3 &  \rb 59.3 & \rb 43.2 & \rb 17.3 & \rb 27,507 & \rb 117,477 & \rb 3,303 \\ %

                RelationTrack \cite{RelationTrack} & \bb & \bb & \bb &  \bb & \bb & \bb & \bb & \bb & \bb  & \bb& \rb \dc{5d1+CH} & \rb 73.8 & \rb 81.0 & \rb 74.7 &  \rb \underline{\textbf{61.0}} & \rb 41.7& \rb 23.2 & \rb 27,999 & \rb 118,623 & \rb \underline{\textbf{1,374}} \\ %
                 
                CSTrack \cite{CSTrack} & \bb & \bb & \bb &  \bb & \bb & \bb & \bb & \bb & \bb  & \bb & \rb\dc{5d1+CH}& \rb74.9 &  \rb80.9 & \rb 72.6 & \rb 59.3 & \rb 41.5 & \rb 17.5 & \rb \underline{\textbf{23,847}}	& \rb 114,303 & \rb 3,567 \\ %
                
                MLT \cite{MLT_Zhang2020} & \bb & \bb & \bb &  \bb & \bb & \bb & \bb & \bb & \bb  & \bb & \rb(\dc{5d1+CH}) & \rb 75.3  & \rb\underline{\textbf{81.7}} & \rb\underline{\textbf{75.5}} & \bb & \rb 49.3 & \rb 19.5 & \rb {27,879} & \rb 109,836 & \rb 1,719  \\ %
                
                FUFET \cite{FUFET} &\bb& \bb & \bb & \bb &\bb&\bb& \bb &\bb & \bb 
                 & \bb & \rb(\dc{5d1+CH})& \rb 76.2 &  \rb81.1 &  \rb68.0 & \rb 57.9 & \rb{{51.1}} &\rb 13.6 &\rb 32,796 &	\rb{98,475} & \rb3,237 \\
                
                TransCenterV2 \cite{Transcenter_Xu2022}   &  \rb\dc{5d1+CH} & \rb \underline{\textbf{76.0}} & \rb \underline{\textbf{81.4}} & \rb 65.6 & \bb & \rb \underline{\textbf{47.3}} & \rb \underline{\textbf{15.3}}  & \rb  28,369  & \rb \underline{\textbf{101,988}} & \rb 4,972    & \rb \dc{5d1+CH} & \rb 76.4 & \rb 81.2 &\rb 65.4 & \bb &\rb \underline{\textbf{51.7}} & \rb \underline{\textbf{11.6}} & \rb 37,005  & \rb \underline{\textbf{89,712}} & \rb 6,402 \\
                
                \midrule
                MOTDT17 \cite{MOTDT17} & \ob\dc{re1}& \ob50.9 & \ob76.6 & \ob52.7 & \ob 41.2 & \ob17.5 & \ob35.7 & \ob24,069 & \ob250,768 & \ob2,474  & \bb & \bb & \bb &  \bb & \bb & \bb & \bb & \bb & \bb  & \bb \\
                 
                UnsupTrack \cite{UnsupTrack} &  \ob\dc{pt}& \ob61.7  & \ob78.3 & \ob58.1 & \ob 46.9 & \ob27.2 & \ob32.4 & \ob 16,872 & \ob197,632 & \ob 1,864 
                  & \bb & \bb & \bb &  \bb & \bb & \bb & \bb & \bb & \bb  & \bb \\ %
                 
                GMT\_CT \cite{GMT_CT2021} & \ob\dc{re2}& \ob 61.5 & \bb & \ob 66.9 & \bb & \ob 26.3 & \ob 32.1 & \ob \underline{\textbf{14,059}} & \ob 200,655 & \ob 2,415 & \bb & \bb & \bb &  \bb & \bb & \bb & \bb & \bb & \bb  & \bb\\

                TrackFormer \cite{Trackformer} &  \ob\dc{CH} & \ob62.5 & \bb & \ob60.7 & 
                \bb & \ob 29.8 & \ob{26.9} & \ob14,966 & \ob206,619 & \ob \underline{\textbf{1,189}}  & \ob\dc{CH} & \ob 74.1 & \bb &  \ob 68.0 & \ob 57.3 & \ob 47.3 & \ob 10.4 & \ob 34,602  & \ob 108,777 & \ob 2,829  \\ %
                
                SiamMOT \cite{SiamMOT} & \ob\dc{ch} & \ob 65.9 & \bb & \ob 63.5 & \bb & \ob 34.6 & \ob 23.9 & \ob 18,098 & \ob 170,955 & \ob 3,040 
                   & \bb & \bb & \bb &  \bb & \bb & \bb & \bb & \bb & \bb  & \bb \\
                 
                MOTR \cite{MOTR_Zeng2021} & \ob\dc{CH} & \ob 67.4 & \bb & \ob 67.0 & \bb & \ob 34.6 & \ob 24.5 & \ob 32,355 & \ob 149,400 & \ob 1,992 
                   & \ob\dc{CH}& \ob73.4 & \bb & \ob 68.6 & \ob 57.8 & \bb & \bb & \bb & \bb & \ob \underline{\textbf{2439}}  \\
                 
                 CenterTrack \cite{centertrack} & \bb& \bb & \bb & \bb & \bb  & \bb  & \bb & \bb & \bb
                 &\bb  & \ob\dc{CH}& \ob67.8 &  \ob78.4 & \ob 64.7 & \bb & \ob34.6&  \ob24.6&  \ob \underline{\textbf{18,489}} & \ob160,332 & \ob 3,039 
                  \\
                
                TraDeS \cite{TraDeS} & \bb & \bb & \bb &  \bb & \bb & \bb & \bb & \bb & \bb  & \bb &  \ob\dc{CH}& \ob69.1 & \bb & \ob63.9 & \ob 52.7 & \ob36.4 & \ob21.5 & \ob 20,892 & \ob150,060 & \ob 3,555 \\ %
                 
                PermaTrack \cite{PermaTrack} & \bb & \bb & \bb &  \bb & \bb & \bb & \bb & \bb & \bb  & \bb& \ob\dc{CH} & \ob 73.8 & \bb & \ob 68.9 & \ob 55.5 & \ob 43.8  & \ob 17.2  & \ob 28,998 & \ob 114,104  & \ob 3,699 \\
                  
                TransTrack \cite{TransTrack_Sun2020} & \bb & \bb & \bb &  \bb & \bb & \bb & \bb & \bb & \bb  & \bb&  \ob\dc{CH}& \ob75.2 & \bb & \ob 63.5 & \ob 54.1 & \ob 55.3 & \ob 10.2 & \ob 50,157 & \ob \underline{\textbf{86,442}} & \ob 3,603  
                 \\
                 
                TransCenterV2 \cite{Transcenter_Xu2022} &  \ob\dc{CH}  & \ob \underline{\textbf{75.9}}  & \ob 81.2 & \ob 65.9  & \ob 56.7 & \ob 49.8  & \ob 12.1  & \ob 30,190  & \ob 100,999 & \ob  4,626 
                 & \ob \dc{CH} &\ob \underline{\textbf{76.2}} & \ob \underline{\textbf{81.1}} &\ob 65.5 & \ob56.7 &\ob 53.5  & \ob 7.9 & 40,101\ob   & \ob 88,827 & \ob 5,394   \\
                
                  \midrule
                
                STC (ours)  &  \ob\dc{CH}  & \ob \underline{\textbf{75.9}}  & \ob \underline{\textbf{81.2}} & \ob \underline{\textbf{72.8}} & \ob \underline{\textbf{60.2}} & \ob \underline{\textbf{50.2}} & \ob \underline{\textbf{11.6}} & \ob 34,615  & \ob \underline{\textbf{98,617}} & \ob 2,860 
                 & \ob \dc{CH} & \ob 75.9 & \ob 81.0 &\ob \underline{\textbf{72.7}} & \ob \underline{\textbf{60.3}} & \ob \underline{\textbf{54.3}} & \ob \underline{\textbf{7.3}} & 45,939\ob   & \ob 86,748 & \ob 3,177  \\

            \bottomrule
            \end{tabular}
    }
\end{center}    
\end{table*}

\begin{table*}[htbp]
    \tabcolsep=0.11cm
    \caption{Results on MOT20 testset on public and private detections. The methods chosen for comparison use the same dataset for pretraining or additional datasets, indicated by the background color. The best result within the same training conditions (background color) is in \textbf{bold}.} \label{table:mot20}
    \resizebox{\linewidth}{!}{
            \begin{tabular}{ l | c c c c c c c c c c || c c c c c c c c c c}
            \toprule
            & \multicolumn{10}{c||}{Public Detections} & \multicolumn{10}{c}{Private Detections}\\\midrule
              Method & Data &  MOTA $\uparrow$ & MOTP $\uparrow$ & IDF1 $\uparrow$ & HOTA $\uparrow$ & MT $\uparrow$ & ML $\downarrow$ & FP $\downarrow$ & FN $\downarrow$ & IDSW $\downarrow$  & Data &  MOTA $\uparrow$ & MOTP $\uparrow$ & IDF1 $\uparrow$ & HOTA $\uparrow$ & MT $\uparrow$ & ML $\downarrow$ & FP $\downarrow$ & FN $\downarrow$ & IDSW $\downarrow$  \\\midrule
            
            CorrTracker \cite{CorrTracker}& \bb & \bb & \bb &  \bb & \bb & \bb & \bb & \bb & \bb & \bb& \rb \dc{5d1} & \rb 65.2 & \bb & \rb 69.1 & \bb & \rb {66.4} & \rb {8.9} & \rb 79,429 & \rb 95,855 & \rb 5,183
             \\ %

            GSDT\_V2 \cite{GSDT} &\bb&\bb&\bb&\bb&\bb&\bb&\bb&\bb&\bb&\bb& \rb\dc{5d1}& \rb 67.1 & \bb &  \rb67.5 & \rb 53.6 & \rb53.1 & \rb13.2 & \rb{31,507} &\rb 135,395 & \rb 3,230 \\ %
            
            GSDT \cite{GSDT} &\bb&\bb&\bb&\bb&\bb&\bb&\bb&\bb&\bb&\bb& \rb\dc{5d1}& \rb67.1 &  \rb{79.1} &  \rb67.5 & \rb 53.6 & \rb53.1 & \rb13.2 & \rb31,913 &	\rb135,409	 & \rb{{3,131}} \\ %
            
            SOTMOT \cite{SOTMOT} &\bb&\bb&\bb&\bb&\bb&\bb&\bb&\bb&\bb&\bb & \rb \dc{5d1} & \rb {{68.6}} & \bb & \rb \underline{\textbf{71.4}} & \rb \underline{\textbf{57.4}} & \rb 64.9 & \rb 9.7 & \rb 57,064 & \rb 101,154 & \rb 4,209 \\

            FairMOT \cite{FairMOT} &\bb&\bb&\bb&\bb&\bb&\bb&\bb&\bb&\bb& \bb& \rb\dc{5d1+CH}& \rb61.8 & \rb78.6 &  \rb 67.3  & \rb 54.6 & \rb\underline{\textbf{68.8}}  & \rb\underline{\textbf{7.6}}  & \rb103,440& \rb \underline{\textbf{88,901}} & \rb5,243   \\  %
            
            CSTrack \cite{CSTrack} &\bb&\bb&\bb&\bb&\bb&\bb&\bb&\bb&\bb&\bb&  \rb\dc{5d1+CH}& \rb66.6 &  \rb78.8 &  \rb68.6  & \rb 54.0 & \rb50.4 & \rb15.5 &\rb\underline{\textbf{25,404}} &\rb144,358 & \rb3,196 \\ %
             
            RelationTrack \cite{RelationTrack} & \bb & \bb & \bb &  \bb & \bb & \bb & \bb & \bb & \bb  & \bb& \rb \dc{5d1+CH} & \rb 67.2 & \rb 79.2 & \rb 70.5 &  \rb 56.5 & \rb 62.2& \rb 8.9 & \rb 61,134 & \rb 104,597 & \rb 4,243 \\ %
            
            TransCenterV2 \cite{Transcenter_Xu2022}  &  \rb\dc{5d1+CH} & \rb \underline{\textbf{72.4}} &\rb \underline{\textbf{81.2}} & \rb \underline{\textbf{57.9}} & \bb &\rb \underline{\textbf{64.2}} & \rb \underline{\textbf{12.3}} & \rb \underline{\textbf{25,121}}   & \rb \underline{\textbf{115,421}} & \rb  \underline{\textbf{2,290}}
            &  \rb \dc{5d1+CH} & \rb \underline{\textbf{72.5}} & \rb \underline{\textbf{81.1}} & \rb 58.1  & \bb & \rb 64.7 & \rb 12.2  & \rb 25,722 & \rb  114,310 & \rb \underline{\textbf{2,332}}  \\
            
            \midrule 
            UnsupTrack \cite{UnsupTrack} & \ob\dc{pt}& \ob53.6 & \ob{80.1} & \ob{50.6}  & \ob 41.7 & \ob30.3  & \ob25.0  & \ob\underline{\textbf{6,439}} & \ob231,298& \ob 2,178   &\bb&\bb&\bb&\bb&\bb&\bb&\bb&\bb &\bb& \bb\\
            
            TransTrack \cite{TransTrack_Sun2020} & \bb & \bb & \bb &  \bb & \bb & \bb & \bb & \bb  & \bb&\bb  & \ob\dc{CH}& \ob65.0 & \bb & \ob 50.1 & \ob 48.9 & \ob  & \ob13.4 & \ob 27,191 & \ob150,197 & \ob3,608
              \\ 

            Trackformer \cite{Trackformer} & \bb & \bb & \bb &  \bb & \bb & \bb & \bb & \bb  & \bb&\bb  & \ob\dc{CH}& \ob68.6 & \bb  & \ob 65.7 & \ob 54.7 & \ob53.6& \ob14.6& \ob \underline{\textbf{20,348}} & \ob 140,373 & \ob \underline{\textbf{1,532}}\\ 
            
            TransCenterV2 \cite{Transcenter_Xu2022} &  \ob\dc{CH} & \ob 72.8 & \ob \underline{\textbf{81.0}}  &\ob 57.6 & \ob 50.1 &\ob 65.5  & \ob 12.1 & \ob 28,026  & \ob \underline{\textbf{110,312}} & \ob 2,621
            &  \ob \dc{CH} & \ob \underline{\textbf{72.9}} & \ob \underline{\textbf{81.0}} & \ob  57.7 & \ob 50.2 & \ob 66.5 & \ob \underline{\textbf{11.8}} & \ob 28,596 & \ob \underline{\textbf{108,982}} & \ob 2,625  \\

            \midrule
            
            STC (Ours)   &  \ob\dc{CH} & \ob \underline{\textbf{72.8}} &\ob \underline{\textbf{81.0}} & \ob \underline{\textbf{68.1}} & \ob \underline{\textbf{56.4}} &\ob \underline{\textbf{65.9}} & \ob \underline{\textbf{12.1}} & \ob 28,167  & \ob 110,687 & \ob  \underline{\textbf{1,769}}
            &  \ob \dc{CH} & \ob \underline{\textbf{72.9}} & \ob \underline{\textbf{81.0}} & \ob \underline{\textbf{68.2}} & \ob \underline{\textbf{56.4}} & \ob \underline{\textbf{66.8}} & \ob  11.9  & \ob 28,892 & \ob  109,309 & \ob 1,791 \\
            \bottomrule
            \end{tabular}
    }
\end{table*}

\subsection{Offline modules} \label{offline}
The tracker we presented is an \textbf{online} tracker (causal), which means it generates the prediction for the track locations on each frame only based on information from the current and previous frames. Other trackers \cite{ReMOT_Yang2021,StrongSORT,SG_offline_Zhen2012} developed methods for offline tracking (non-causal), which have the advantage of using information from future frames as well. These trackers can be used for various tasks such as sports analysis or analysis of past surveillance videos, but not for real-time tasks which require immediate response.
Nevertheless, we decided to test two non-causal features in our ablation study \ref{ablation_study} - AFLink and GSI, both of which were developed in the work of StrongSORT \cite{StrongSORT}.\\
\textbf{AFLink} (Appearance-Free Link) is a linking algorithm, predicting connectivity between two trajectories based on the spatio-temporal information, i.e., the change of their respective positions through time. It is essentially able to conclude whether two predicted tracks by the original tracker are in fact from the same object and therefore decrease the number of IDSWs.\\
\textbf{GSI} (Gaussian-smoothed Interpolation) is an interpolation algorithm, intended to fill gaps in predicted trajectories by the original tracker. It models the trajectory as a gaussian process with a function kernel in order to interpolate the track's position in the missing segment.

\subsection{Ablation Study} \label{ablation_study}
In our ablation study we compared the added modules described in \ref{proposed_method} that make up our proposed method STC, as well as the offline modules described in \ref{offline} that we explored on top of STC and the original TransCenter \cite{Transcenter_Xu2022}. The AFLink and GSI modules are the non-causal features developed in \cite{StrongSORT} and described in \ref{offline}. As we can see in the results Table III the AFLink module mainly improves the IDSW number and has no significant effect on the main metrics. The GSI module has a significant effect on the number of FP (false positives) and FN (false negatives, i.e., misdetections) since it was able to complete missing trajectories but at the same time generated false trajectories, and the balance between them indeed improved the main evaluation metrics in several configurations, depending on parameters of the GSI function.
In order to optimize the association step between tracks and new detections we tested association using only the appearance embeddings (noted as "Only Embedding"), with the addition of the default implementation of the Kalman filter based on DeepSORT \cite{Wojke2017} (noted as "Default Kalman") and with the addition of our improved implementation of the Kalman filter (noted as "Improved Kalman"). The improved Kalman and the embedding network together make up our STC tracker. Using only the embedding distance caused a very high number of IDSW and poor metric results as expected, since it caused the association to match tracks and detections with no regard to their positions. Using only the improved Kalman filter, without using the embedding distance, i.e., not using the appearance feature at all, caused a slightly smaller number of FP but higher number of IDSW. We can also see that the improved Kalman filter contributes to all the metrics compared to the default Kalman implementation, which confirms that modifying the state vector, correcting the measurement projection matrix and adapting the measurement noise covariance matrix using the detection heatmap FWHM are an effective way to fine-tune the tracker to the required task. The last experiment is testing the classic method of fusing the embedding and GIoU scores for the final association distance, using the formula: 
\begin{equation}
    Dist = \lambda \cdot d_{emb} + (1 - \lambda)\cdot d_{giou} 
\label{fuse_dist_formula}
\end{equation}
Where $Dist$ is the fused association distance, $d_{emb}$ is the appearance embedding distance, $d_{giou}$ is the bounding box GIoU distance and $\lambda$ is a configurable parameter for the respective weight, that we chose as $0.5$ for this example after testing various values. We found that like in \cite{Aharon2022} the best way to utilize the information from the embedding distances is by using the minimum between the two distances. The minimum is taken after filtering the GIoU distance with a higher threshold to guarantee the track predicted position and the positions of the potential matches are close, before considering a match based on appearance.
\begin{table*}[htbp]
\begin{center}
    \caption{Ablation study of different components of our tracker. The experiments were conducted on the mot17 train dataset using private detections. The rows in {\color{blue}dark blue} text are offline options. Best results are \underline{underlined}, best online results are in \textbf{bold}.}
    \label{table:ablation_table}
    \resizebox{\linewidth}{!}{
        \begin{tabular}{=c|+c|+c|+c|+c|+c|+c}
        \toprule
        Setting & MOTA $\uparrow$ & IDF1 $\uparrow$ & HOTA $\uparrow$ 
            & FP $\downarrow$ & FN $\downarrow$ & IDSW $\downarrow$ \\
        \midrule
            TransCenterV2 & \textbf{86.9} & 77.3 & 70.4 & \textbf{1272} & 12813 & 678 \\
            \rowstyle{\color{blue}}
            TransCenterV2 + AFLink & 86.9 & 79.9 & 71.7& \underline{1251} & 12892 & 603 \\
            \rowstyle{\color{blue}}
            TransCenterV2 + GSI & 87.0 & 77.2 & 70.4 & 2442 & 11490 & 663 \\ 
            \rowstyle{\color{blue}}
            TransCenterV2 + AFLink + GSI & 87.1 & 79.9 & 71.9 & 2306 & 11583 & 565 \\
            TransCenterV2 + Default Kalman & 86.7 & 82.0 & 73.0 & 1553 & 12779 & 622 \\
            TransCenterV2 + Improved Kalman & 86.8 & 82.7 & 73.4 & 1539 & 12783 & 524  \\
            TransCenterV2 + Default Kalman + Embedding & 86.7 & 82.8 & 73.6 & 1611 & 12802 & 583  \\
            \textbf{STC} & 86.8 & \textbf{83.6} & \textbf{73.9} & 1581 & \textbf{12750} & \textbf{453}  \\
            \rowstyle{\color{blue}}
            STC + AFLink & 86.8 & 83.1 & 73.6 & 1563 & 12798 & 436  \\
            \rowstyle{\color{blue}}
            STC + GSI & \underline{88.3} & \underline{84.0} & \underline{74.6} & 2057 & \underline{10782} & 329  \\
            \rowstyle{\color{blue}}
            STC + AFLink + GSI & 88.2 & 83.6 & 74.4 & 2145 & 10784 & \underline{305} \\
            STC Only {Embedding} & 85.7 & 76.2 & 69.5 & 1435 & 12767 & 1820 \\
            STC {Fused Embedding \& GIoU} & 86.7 & {80.5} & {72.3} & 1325 & 12789 & 624 \\
            
            \bottomrule
        \end{tabular}
    }
\end{center}
    
\end{table*}

\subsection{Qualitative Analysis} \label{qualitative_analysis}
\begin{figure}[htpb]
        \centering
        \includegraphics[width=0.9\linewidth]{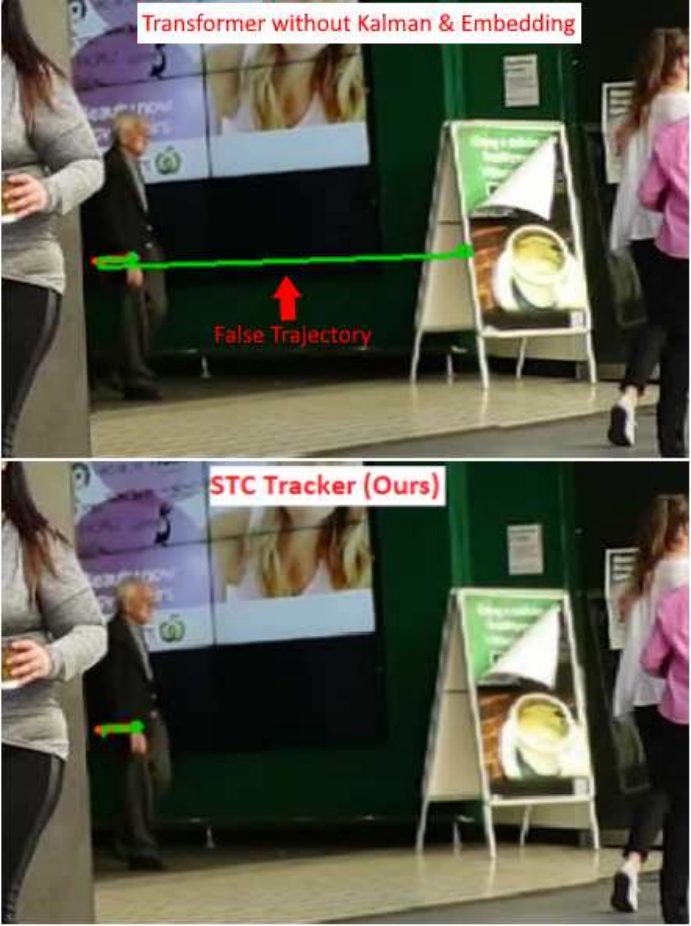}
        \caption{Visualization of a specific scenario in the results of a tracker based only on transformer (TransCenterV2 \cite{Transcenter_Xu2022}) at the top and our STC tracker results (Transformer with Kalman and Embedding) at the bottom. The green trajectories are the tracker prediction with a history of 20 frames and the orange trajectories are the ground truth positions. An IDSW has occurred in the top image, and is demonstrated by the big "jump" in the trajectory from right to left, giving the new pedestrian that emerged an existing track ID. In the bottom image the trajectory only began when the pedestrian emerged. The results are from the MOT17-09 video on frame 389. }
        \label{fig:idsw_example1}
\end{figure}
\begin{figure}[htb]
    \centering
    \hspace*{-1mm}     
    \includegraphics[width=1\linewidth]{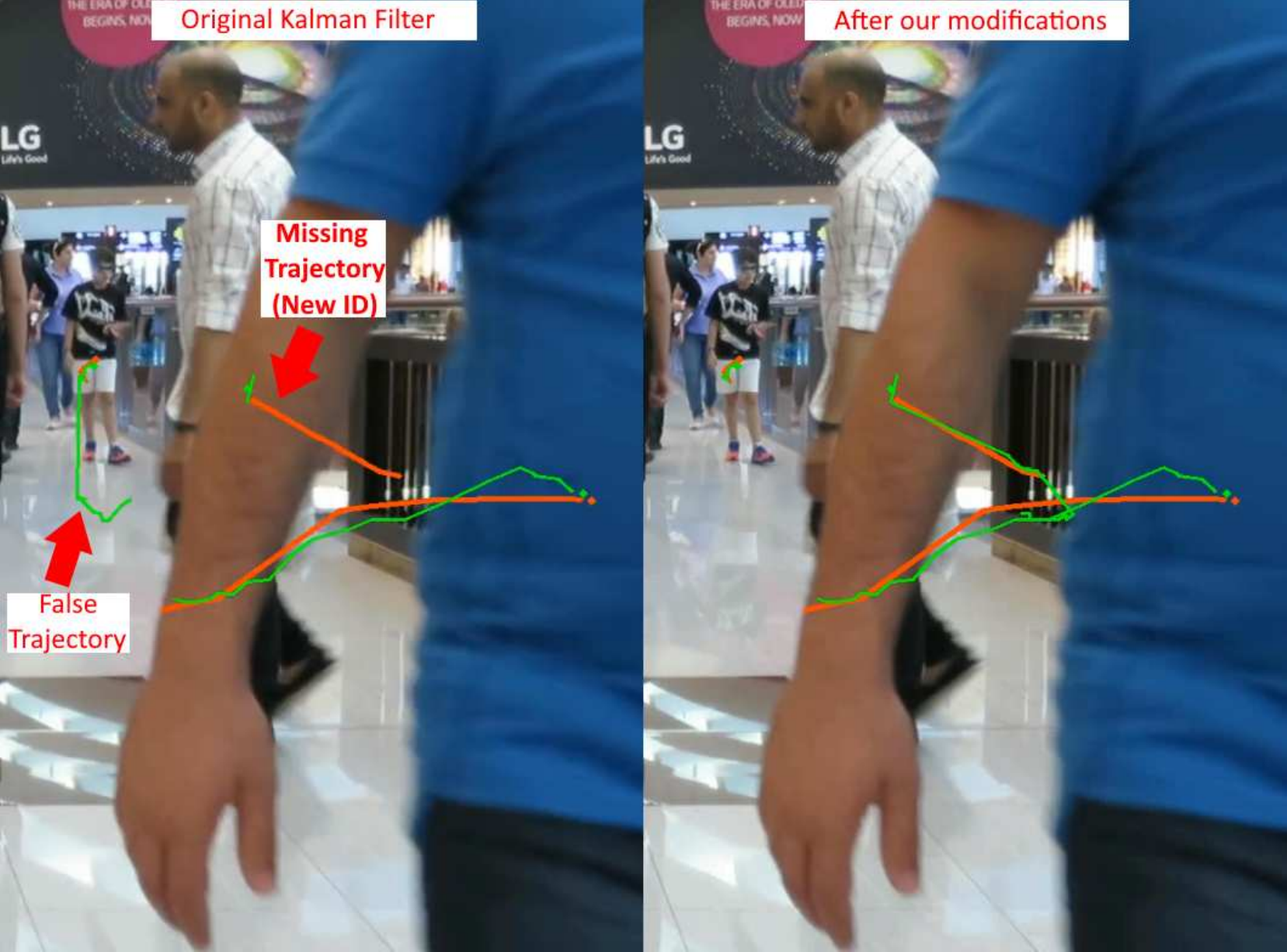}
    \caption{Visualization of a specific scenario in the results of our tracker when based on the default Kalman implementation seen in many works \cite{Wojke2017,StrongSORT,ByteTrack_Zhang2021,FairMOT}, compared with our tracker after the modifications to the Kalman filter. The green trajectories are the tracker prediction with a history of 20 frames and the orange trajectories are the ground truth positions. The image on the left shows a person reappearing from the right after an occlusion and receiving a new ID by the tracker, while a new person in the back appears in the frame and receives a false existing ID with a long history. In the image on the right, both errors are corrected with the modified Kalman filter. The results are from the MOT17-11 video on frame 720. }
    \label{fig:idsw_example2}
\end{figure}
\begin{figure*}[htp]
    \begin{center}    
    \includegraphics[width=0.95\textwidth]{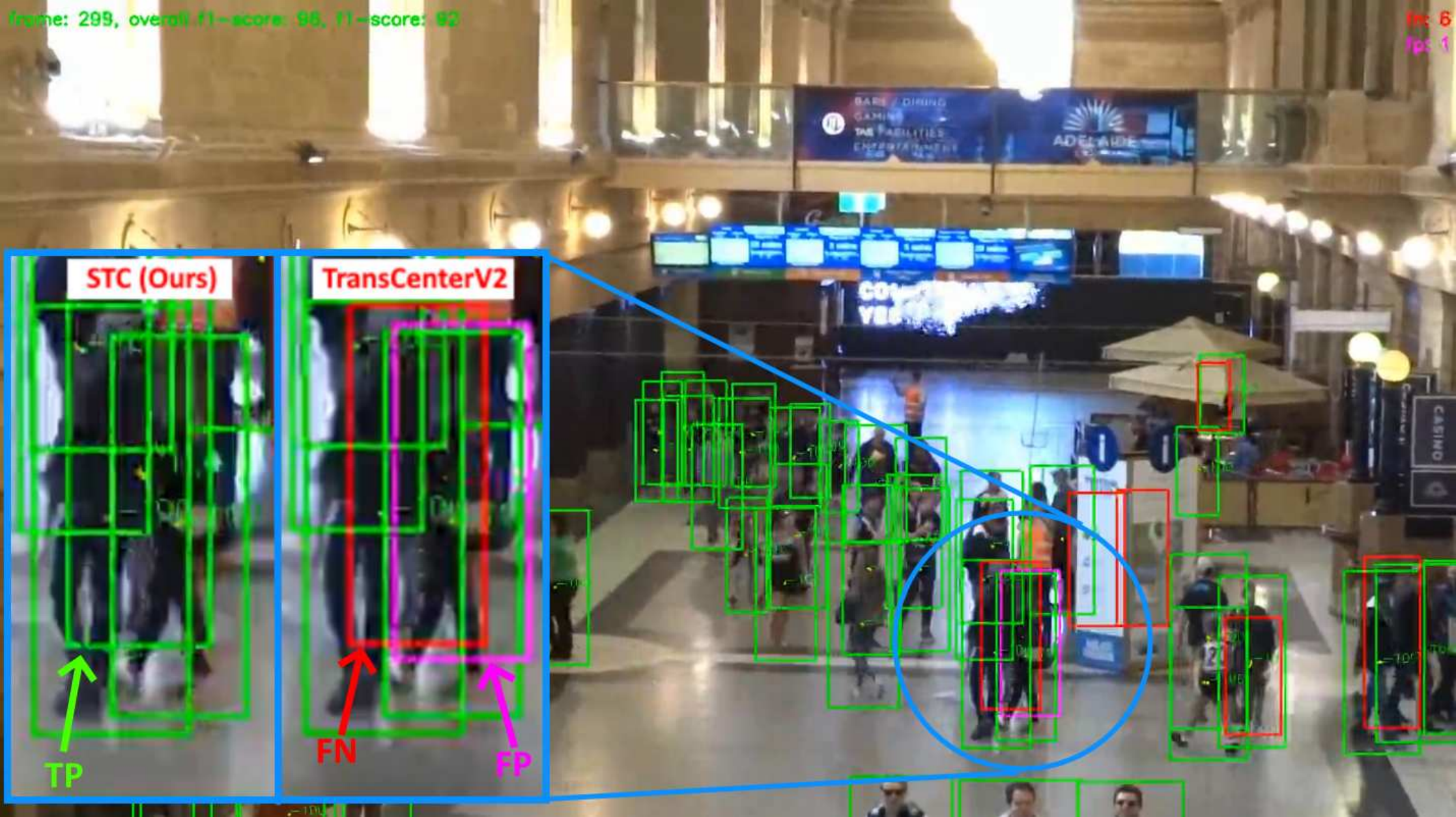}
    \end{center}
    \caption{Visualization of a specific scenario in the results of a tracker based only on transformer (TransCenterV2 \cite{Transcenter_Xu2022}). The full frame shows the results of the TransCenterV2 tracker compared to the ground truth: green boxes are True Positive (TP), red boxes are False negative (FN) and pink boxes are False Positive (FP). The zoom-in image shows the area of the error in the frame with the original TransCenter on the right and with our STC tracker on the left. The results are from the MOT20-01 video on frame 299. }
    \label{fig:fn_example1}
\end{figure*}

Another method of analysis included plotting bounding boxes to indicate true positive tracking (TP) in green, false negative (FN) in red and false positive (FP) in pink. An additional example for improved tracking is demonstrated in Fig \ref{fig:fn_example1}, in which the indicated boxes show a pedestrian that is almost entirely occluded and was detected by the Transformer attention mechanism in the wrong position, but the addition of a motion model made it possible to associate it to the existing track in our STC tracker. The original TransCenter managed to detect the mostly occluded pedestrian but positioned the track in the wrong location, causing both an FN in the ground truth location and an FP in the tracker predicted location. Our STC tracker managed to predict the correct location in this scenario due to the added Kalman-based motion model. This demonstrates both the ability of the transformer attention mechanism to infer the existence of occluded objects and the assistance of the motion model in predicting the location of lost objects. 

\section{Conclusion} \label{conclusion}
In this paper, we explored the problem of multi-object tracking and demonstrated the potential of a Transformer-based tracker on the MOTChallenge MOT17 and MOT20 benchmark datasets. We addressed one of the weakest components of the transformer - its motion estimation branch - and demonstrated the benefits of including a fine-tuned Kalman filter for motion estimation while choosing a more accurate representation of the measurement vector and estimating its error. Moreover, we found that integrating a combination of appearance embedding and position in the track association stage  reduces association errors as demonstrated by the IDF1 metric. Our tracker is currently ranked first among the transformer-based trackers in the MOTChallenge datasets in terms of HOTA and IDF1, indicating the merits of our modifications. Looking ahead, we believe future research into transformers and their tailored implementation for the MOT task might lead to an all-in-one transformer based tracker without the need for additional modules.

\section*{Acknowledgement}
We thank Shlomo Shmeltzer Institute for Smart Transportation at Tel-Aviv University for the scholarship for the first author and for the support of our Autonomous Mobile Laboratory.



\printbibliography 

@inproceedings{SG_offline_Zhen2012,
   author = {Zhen Qin and C. R. Shelton},
   doi = {10.1109/CVPR.2012.6247899},
   crossref = {proc-cvpr-2012},
   isbn = {978-1-4673-1228-8},
   journal = {2012 IEEE Conference on Computer Vision and Pattern Recognition},
   month = {6},
   pages = {1972-1978},
   publisher = {IEEE},
   title = {Improving multi-target tracking via social grouping},
   year = {2012},
}

@article{ReMOT_Yang2021,
   author = {Fan Yang and Xin Chang and Sakriani Sakti and Yang Wu and Satoshi Nakamura},
   doi = {10.1016/j.imavis.2020.104091},
   issn = {02628856},
   journal = {Image and Vision Computing},
   month = {2},
   pages = {104091},
   title = {ReMOT: A model-agnostic refinement for multiple object tracking},
   volume = {106},
   year = {2021},
}

@book{Kalman_book,
   author = {Robert Grover Brown and Patrick Y.C. Hwang},
   edition = {4th},
   isbn = {978-0-470-60969-9},
   publisher = {Wiley},
   title = {Introduction to random signals and applied Kalman Filtering},
   year = {2012},
}

@inproceedings{MOTDT17,
   author = {Long Chen and Haizhou Ai and Zijie Zhuang and Chong Shang},
   doi = {10.1109/ICME.2018.8486597},
   crossref = {proc-icme-2018},
   isbn = {978-1-5386-1737-3},
   journal = {2018 IEEE International Conference on Multimedia and Expo (ICME)},
   month = {7},
   pages = {1-6},
   publisher = {IEEE},
   title = {Real-Time Multiple People Tracking with Deeply Learned Candidate Selection and Person Re-Identification},
   year = {2018},
}

@inproceedings{IDF1,
   author = {Francesco
and Zou Roger
and Cucchiara Rita
and Tomasi Carlo Ristani Ergys
and Solera},
   city = {Cham},
   editor = {Hervé Hua Gang
and Jégou},
   isbn = {978-3-319-48881-3},
   journal = {Computer Vision – ECCV 2016 Workshops},
   pages = {17-35},
   publisher = {Springer International Publishing},
   title = {Performance Measures and a Data Set for Multi-target, Multi-camera Tracking},
   year = {2016},
}

@inproceedings{TMOH_2021,
   author = {Daniel Stadler and Jurgen Beyerer},
   doi = {10.1109/CVPR46437.2021.01081},
   crossref = {proc-cvpr-2021},
   isbn = {978-1-6654-4509-2},
   journal = {2021 IEEE/CVF Conference on Computer Vision and Pattern Recognition (CVPR)},
   month = {6},
   pages = {10953-10962},
   publisher = {IEEE},
   title = {Improving Multiple Pedestrian Tracking by Track Management and Occlusion Handling},
   year = {2021},
}

@article{PVT_V2,
   author = {Wenhai Wang and Enze Xie and Xiang Li and Deng-Ping Fan and Kaitao Song and Ding Liang and Tong Lu and Ping Luo and Ling Shao},
   doi = {10.1007/s41095-022-0274-8},
   issn = {2096-0433},
   issue = {3},
   journal = {Computational Visual Media},
   month = {9},
   pages = {415-424},
   title = {PVT v2: Improved baselines with Pyramid Vision Transformer},
   volume = {8},
   year = {2022},
}

@inproceedings{GIOU_Rezatofighi_2018_CVPR,
   author = {Hamid Rezatofighi and Nathan Tsoi and JunYoung Gwak and Amir Sadeghian and Ian Reid and Silvio Savarese},
   doi = {10.1109/CVPR.2019.00075},
   crossref = {proc-cvpr-2019},
   isbn = {978-1-7281-3293-8},
   journal = {2019 IEEE/CVF Conference on Computer Vision and Pattern Recognition (CVPR)},
   month = {6},
   pages = {658-666},
   publisher = {IEEE},
   title = {Generalized Intersection Over Union: A Metric and a Loss for Bounding Box Regression},
   year = {2019},
}

@article{MOT20_Dendorfer2020,
   author = {Patrick Dendorfer and Hamid Rezatofighi and Anton Milan and Javen Shi and Daniel Cremers and Ian Reid and Stefan Roth and Konrad Schindler and Laura Leal-Taixé},
   journal = {arXiv},
   month = {3},
   title = {MOT20: A benchmark for multi object tracking in crowded scenes},
   year = {2020},
}

@article{FUFET,
   author = {Chaobing Shan and Chunbo Wei and Bing Deng and Jianqiang Huang and Xian-Sheng Hua and Xiaoliang Cheng and Kewei Liang},
   journal = {arXiv},
   month = {10},
   title = {Tracklets Predicting Based Adaptive Graph Tracking},
   year = {2020},
}

@article{MLT_Zhang2020,
   author = {Yang Zhang and Hao Sheng and Yubin Wu and Shuai Wang and Wei Ke and Zhang Xiong},
   doi = {10.1109/JIOT.2020.2996609},
   issn = {2327-4662},
   issue = {9},
   journal = {IEEE Internet of Things Journal},
   month = {9},
   pages = {7892-7902},
   title = {Multiplex Labeling Graph for Near-Online Tracking in Crowded Scenes},
   volume = {7},
   year = {2020},
}

@article{CSTrack,
   author = {Chao Liang and Zhipeng Zhang and Xue Zhou and Bing Li and Shuyuan Zhu and Weiming Hu},
   doi = {10.1109/TIP.2022.3165376},
   issn = {1057-7149},
   journal = {IEEE Transactions on Image Processing},
   pages = {3182-3196},
   title = {Rethinking the Competition Between Detection and ReID in Multiobject Tracking},
   volume = {31},
   year = {2022},
}

@article{RelationTrack,
   author = {En Yu and Zhuoling Li and Shoudong Han and Hongwei Wang},
   doi = {10.1109/TMM.2022.3150169},
   issn = {1520-9210},
   journal = {IEEE Transactions on Multimedia},
   pages = {1-1},
   title = {RelationTrack: Relation-aware Multiple Object Tracking with Decoupled Representation},
   year = {2022},
}

@inproceedings{CorrTracker,
   author = {Qiang Wang and Yun Zheng and Pan Pan and Yinghui Xu},
   doi = {10.1109/CVPR46437.2021.00387},
   crossref = {proc-cvpr-2021},
   isbn = {978-1-6654-4509-2},
   journal = {2021 IEEE/CVF Conference on Computer Vision and Pattern Recognition (CVPR)},
   month = {6},
   pages = {3875-3885},
   publisher = {IEEE},
   title = {Multiple Object Tracking with Correlation Learning},
   year = {2021},
}

@inproceedings{SOTMOT,
   author = {Linyu Zheng and Ming Tang and Yingying Chen and Guibo Zhu and Jinqiao Wang and Hanqing Lu},
   doi = {10.1109/CVPR46437.2021.00248},
   crossref = {proc-cvpr-2021},
   isbn = {978-1-6654-4509-2},
   journal = {2021 IEEE/CVF Conference on Computer Vision and Pattern Recognition (CVPR)},
   month = {6},
   pages = {2453-2462},
   publisher = {IEEE},
   title = {Improving Multiple Object Tracking with Single Object Tracking},
   year = {2021},
}

@inproceedings{GSDT,
   author = {Yongxin Wang and Kris Kitani and Xinshuo Weng},
   doi = {10.1109/ICRA48506.2021.9561110},
   crossref = {proc-icra-2021},
   isbn = {978-1-7281-9077-8},
   journal = {2021 IEEE International Conference on Robotics and Automation (ICRA)},
   month = {5},
   pages = {13708-13715},
   publisher = {IEEE},
   title = {Joint Object Detection and Multi-Object Tracking with Graph Neural Networks},
   year = {2021},
}

@inproceedings{PermaTrack,
   author = {Pavel Tokmakov and Jie Li and Wolfram Burgard and Adrien Gaidon},
   doi = {10.1109/ICCV48922.2021.01068},
   crossref = {proc-iccv-2021},
   isbn = {978-1-6654-2812-5},
   journal = {2021 IEEE/CVF International Conference on Computer Vision (ICCV)},
   month = {10},
   pages = {10840-10849},
   publisher = {IEEE},
   title = {Learning to Track with Object Permanence},
   year = {2021},
}

@inproceedings{TraDeS,
   author = {Jialian Wu and Jiale Cao and Liangchen Song and Yu Wang and Ming Yang and Junsong Yuan},
   doi = {10.1109/CVPR46437.2021.01217},
   crossref = {proc-cvpr-2021},
   isbn = {978-1-6654-4509-2},
   journal = {2021 IEEE/CVF Conference on Computer Vision and Pattern Recognition (CVPR)},
   month = {6},
   pages = {12347-12356},
   publisher = {IEEE},
   title = {Track to Detect and Segment: An Online Multi-Object Tracker},
   year = {2021},
}

@inproceedings{centertrack,
author="Zhou, Xingyi
and Koltun, Vladlen
and Kr{\"a}henb{\"u}hl, Philipp",
editor="Vedaldi, Andrea
and Bischof, Horst
and Brox, Thomas
and Frahm, Jan-Michael",
title="Tracking Objects as Points",
booktitle="Computer Vision -- ECCV 2020",
year="2020",
publisher="Springer International Publishing",
address="Cham",
pages="474--490",
isbn="978-3-030-58548-8"
}

@inproceedings{SiamMOT,
   author = {Bing Shuai and Andrew Berneshawi and Xinyu Li and Davide Modolo and Joseph Tighe},
   doi = {10.1109/CVPR46437.2021.01219},
   crossref = {proc-cvpr-2021},
   isbn = {978-1-6654-4509-2},
   journal = {2021 IEEE/CVF Conference on Computer Vision and Pattern Recognition (CVPR)},
   month = {6},
   pages = {12367-12377},
   publisher = {IEEE},
   title = {SiamMOT: Siamese Multi-Object Tracking},
   year = {2021},
}

@inproceedings{GMT_CT2021,
   author = {Jiawei He and Zehao Huang and Naiyan Wang and Zhaoxiang Zhang},
   crossref = {proc-cvpr-2021},
   journal = {CVPR},
   month = {3},
   pages = {5299-5309},
   title = {Learnable Graph Matching: Incorporating Graph Partitioning with Deep Feature Learning for Multiple Object Tracking},
   year = {2021},
}

@article{UnsupTrack,
   author = {Shyamgopal Karthik and Ameya Prabhu and Vineet Gandhi},
   journal = {arXiv},
   month = {6},
   title = {Simple Unsupervised Multi-Object Tracking},
   year = {2020},
}

@InProceedings{MOTR_Zeng2021,
author="Zeng, Fangao
and Dong, Bin
and Zhang, Yuang
and Wang, Tiancai
and Zhang, Xiangyu
and Wei, Yichen",
editor="Avidan, Shai
and Brostow, Gabriel
and Ciss{\'e}, Moustapha
and Farinella, Giovanni Maria
and Hassner, Tal",
title="MOTR: End-to-End Multiple-Object Tracking with Transformer",
booktitle="Computer Vision -- ECCV 2022",
year="2022",
publisher="Springer Nature Switzerland",
address="Cham",
pages="659--675",
isbn="978-3-031-19812-0"
}

@inproceedings{ResNet2016,
   author = {Kaiming He and Xiangyu Zhang and Shaoqing Ren and Jian Sun},
   doi = {10.1109/CVPR.2016.90},
   crossref = {proc-cvpr-2016},
   isbn = {978-1-4673-8851-1},
   journal = {2016 IEEE Conference on Computer Vision and Pattern Recognition (CVPR)},
   month = {6},
   pages = {770-778},
   publisher = {IEEE},
   title = {Deep Residual Learning for Image Recognition},
   year = {2016},
}

@article{Milan2017,
   author = {Anton Milan and S. Hamid Rezatofighi and Anthony Dick and Ian Reid and Konrad Schindler},
   doi = {10.1609/aaai.v31i1.11194},
   issn = {2374-3468},
   issue = {1},
   journal = {Proceedings of the AAAI Conference on Artificial Intelligence},
   month = {2},
   title = {Online Multi-Target Tracking Using Recurrent Neural Networks},
   volume = {31},
   year = {2017},
}

@inproceedings{Tracktor_Bergmann2019,
   author = {Philipp Bergmann and Tim Meinhardt and Laura Leal-Taixe},
   crossref = {proc-iccv-2019},
   doi = {10.1109/ICCV.2019.00103},
   isbn = {978-1-7281-4803-8},
   journal = {2019 IEEE/CVF International Conference on Computer Vision (ICCV)},
   month = {10},
   pages = {941-951},
   publisher = {IEEE},
   title = {Tracking Without Bells and Whistles},
   year = {2019},
}

@inproceedings{MAA_Stadler2022,
   author = {Daniel Stadler and Jurgen Beyerer},
   doi = {10.1109/WACVW54805.2022.00019},
   crossref = {proc-wacw-2022},
   isbn = {978-1-6654-5824-5},
   journal = {2022 IEEE/CVF Winter Conference on Applications of Computer Vision Workshops (WACVW)},
   month = {1},
   pages = {133-142},
   publisher = {IEEE},
   title = {Modelling Ambiguous Assignments for Multi-Person Tracking in Crowds},
   year = {2022},
}

@article{CoupledNet2021,
   author = {Tianrui Liu and Wenhan Luo and Lin Ma and Jun-Jie Huang and Tania Stathaki and Tianhong Dai},
   doi = {10.1109/TIP.2020.3038371},
   issn = {1057-7149},
   journal = {IEEE Transactions on Image Processing},
   pages = {754-766},
   title = {Coupled Network for Robust Pedestrian Detection With Gated Multi-Layer Feature Extraction and Deformable Occlusion Handling},
   volume = {30},
   year = {2021},
}

@article{FairMOT,
   author = {Yifu Zhang and Chunyu Wang and Xinggang Wang and Wenjun Zeng and Wenyu Liu},
   doi = {10.1007/s11263-021-01513-4},
   issue = {11},
   journal = {IJCV},
   month = {11},
   pages = {3069-3087},
   title = {FairMOT: On the Fairness of Detection and Re-Identification in Multiple Object Tracking},
   volume = {129},
   year = {2021},
}

@inproceedings{SBS,
   author = {Hao Luo and Youzhi Gu and Xingyu Liao and Shenqi Lai and Wei Jiang},
   crossref = {proc-iccv-2019},
   journal = {CVPR},
   month = {3},
   title = {Bag of Tricks and A Strong Baseline for Deep Person Re-identification},
   year = {2019},
}

@article{ResNeSt,
   author = {Hang Zhang and Chongruo Wu and Zhongyue Zhang and Yi Zhu and Haibin Lin and Zhi Zhang and Yue Sun and Tong He and Jonas Mueller and R. Manmatha and Mu Li and Alexander Smola},
   journal = {CVPR},
   month = {4},
   title = {ResNeSt: Split-Attention Networks},
   year = {2020},
}

@article{FastReID,
   author = {Lingxiao He and Xingyu Liao and Wu Liu and Xinchen Liu and Peng Cheng and Tao Mei},
   journal = {arXiv},
   month = {6},
   title = {FastReID: A Pytorch Toolbox for General Instance Re-identification},
   year = {2020},
}

@article{HOTA,
   author = {Jonathon Luiten and Aljosa Osep and Patrick Dendorfer and Philip Torr and Andreas Geiger and Laura Leal-Taixe and Bastian Leibe},
   doi = {10.1007/s11263-020-01375-2},
   journal = {arXiv},
   month = {9},
   title = {HOTA: A Higher Order Metric for Evaluating Multi-Object Tracking},
   year = {2020},
}

@article{Kuhn1955,
   author = {H. W. Kuhn},
   doi = {10.1002/nav.3800020109},
   issn = {00281441},
   issue = {1-2},
   journal = {Naval Research Logistics Quarterly},
   month = {3},
   pages = {83-97},
   title = {The Hungarian method for the assignment problem},
   volume = {2},
   year = {1955},
}

@article{StrongSORT,
   author = {Yunhao Du and Yang Song and Bo Yang and Yanyun Zhao},
   journal = {arXiv},
   month = {2},
   title = {StrongSORT: Make DeepSORT Great Again},
   year = {2022},
}

@article{Aharon2022,
   author = {Nir Aharon and Roy Orfaig and Ben-Zion Bobrovsky},
   journal = {arXiv},
   month = {6},
   title = {BoT-SORT: Robust Associations Multi-Pedestrian Tracking},
   year = {2022},
}

@article{Attention,
   author = {Ashish Vaswani and Noam Shazeer and Niki Parmar and Jakob Uszkoreit and Llion Jones and Aidan N. Gomez and Lukasz Kaiser and Illia Polosukhin},
   journal = {NeurIPS},
   month = {6},
   pages = {5998-6008},
   title = {Attention Is All You Need},
   year = {2017},
}

@article{Trackformer,
   author = {Tim Meinhardt and Alexander Kirillov and Laura Leal-Taixe and Christoph Feichtenhofer},
   journal = {arXiv},
   month = {1},
   title = {TrackFormer: Multi-Object Tracking with Transformers},
   year = {2021},
}

@article{Kim2022,
  doi = {10.1007/s10489-022-03473-9},
  url = {https://doi.org/10.1007/s10489-022-03473-9},
  year = {2022},
  month = apr,
  publisher = {Springer Science and Business Media {LLC}},
  volume = {53},
  number = {1},
  pages = {930--940},
  author = {Sangwon Kim and Jimi Lee and Byoung Chul Ko},
  title = {{SSL}-{MOT}: self-supervised learning based multi-object tracking},
  journal = {Applied Intelligence}
}

@InProceedings{ByteTrack_Zhang2021,
author="Zhang, Yifu
and Sun, Peize
and Jiang, Yi
and Yu, Dongdong
and Weng, Fucheng
and Yuan, Zehuan
and Luo, Ping
and Liu, Wenyu
and Wang, Xinggang",
editor="Avidan, Shai
and Brostow, Gabriel
and Ciss{\'e}, Moustapha
and Farinella, Giovanni Maria
and Hassner, Tal",
title="ByteTrack: Multi-object Tracking by Associating Every Detection Box",
booktitle="Computer Vision -- ECCV 2022",
year="2022",
publisher="Springer Nature Switzerland",
address="Cham",
pages="1--21",
isbn="978-3-031-20047-2"
}

@article{Luo2021,
   author = {Wenhan Luo and Junliang Xing and Anton Milan and Xiaoqin Zhang and Wei Liu and Tae-Kyun Kim},
   doi = {10.1016/j.artint.2020.103448},
   issn = {00043702},
   journal = {Artificial Intelligence},
   month = {4},
   pages = {103448},
   title = {Multiple object tracking: A literature review},
   volume = {293},
   year = {2021},
}

@article{DETR,
   author = {Nicolas Carion and Francisco Massa and Gabriel Synnaeve and Nicolas Usunier and Alexander Kirillov and Sergey Zagoruyko},
   journal = {arXiv},
   month = {5},
   title = {End-to-End Object Detection with Transformers},
   year = {2020},
}

@article{Transcenter_Xu2022,
   author={Xu, Yihong and Ban, Yutong and Delorme, Guillaume and Gan, Chuang and Rus, Daniela and Alameda-Pineda, Xavier},
  journal={IEEE Transactions on Pattern Analysis and Machine Intelligence}, 
  title={TransCenter: Transformers With Dense Representations for Multiple-Object Tracking}, 
  year={2022},
  volume={},
  number={},
  pages={1-16},
  doi={10.1109/TPAMI.2022.3225078}
}

@article{TransTrack_Sun2020,
   author = {Peize Sun and Jinkun Cao and Yi Jiang and Rufeng Zhang and Enze Xie and Zehuan Yuan and Changhu Wang and Ping Luo},
   journal = {arXiv},
   month = {12},
   title = {TransTrack: Multiple Object Tracking with Transformer},
   year = {2020},
}

@InProceedings{JDE_Wang2019,
author="Wang, Zhongdao
and Zheng, Liang
and Liu, Yixuan
and Li, Yali
and Wang, Shengjin",
editor="Vedaldi, Andrea
and Bischof, Horst
and Brox, Thomas
and Frahm, Jan-Michael",
title="Towards Real-Time Multi-Object Tracking",
booktitle="Computer Vision -- ECCV 2020",
year="2020",
publisher="Springer International Publishing",
address="Cham",
pages="107--122",
isbn="978-3-030-58621-8"
}

@article{MOTA_Bernardin2008,
   author = {Keni Bernardin and Rainer Stiefelhagen},
   doi = {10.1155/2008/246309},
   issn = {1687-5176},
   journal = {EURASIP Journal on Image and Video Processing},
   pages = {1-10},
   title = {Evaluating Multiple Object Tracking Performance: The CLEAR MOT Metrics},
   volume = {2008},
   year = {2008},
}

@inproceedings{Wojke2017,
   author = {Nicolai Wojke and Alex Bewley and Dietrich Paulus},
   doi = {10.1109/ICIP.2017.8296962},
   crossref = {proc-icip-2017},
   isbn = {978-1-5090-2175-8},
   journal = {2017 IEEE International Conference on Image Processing (ICIP)},
   month = {9},
   pages = {3645-3649},
   publisher = {IEEE},
   title = {Simple online and realtime tracking with a deep association metric},
   year = {2017},
}

@inproceedings{Bewley2016,
   author = {Alex Bewley and Zongyuan Ge and Lionel Ott and Fabio Ramos and Ben Upcroft},
   doi = {10.1109/ICIP.2016.7533003},
   crossref = {proc-icip-2016},
   journal = {ICIP},
   month = {2},
   pages = {3464-3468},
   title = {Simple Online and Realtime Tracking},
   year = {2016},
}

@article{Milan2016,
   author = {Anton Milan and Laura Leal-Taixe and Ian Reid and Stefan Roth and Konrad Schindler},
   journal = {arXiv},
   month = {3},
   title = {MOT16: A Benchmark for Multi-Object Tracking},
   year = {2016},
}

@ARTICLE{9229518,
  author={Wang, Huibing and Wang, Yang and Zhang, Zhao and Fu, Xianping and Zhuo, Li and Xu, Mingliang and Wang, Meng},
  journal={IEEE Transactions on Multimedia}, 
  title={Kernelized Multiview Subspace Analysis By Self-Weighted Learning}, 
  year={2021},
  volume={23},
  number={},
  pages={3828-3840},
  doi={10.1109/TMM.2020.3032023}}

@ARTICLE{9106791,
  author={Wang, Huibing and Peng, Jinjia and Chen, Dongyan and Jiang, Guangqi and Zhao, Tongtong and Fu, Xianping},
  journal={IEEE MultiMedia}, 
  title={Attribute-Guided Feature Learning Network for Vehicle Reidentification}, 
  year={2020},
  volume={27},
  number={4},
  pages={112-121},
  doi={10.1109/MMUL.2020.2999464}}

@ARTICLE{10032273,
  author={Wang, Huibing and Yao, Mingze and Jiang, Guangqi and Mi, Zetian and Fu, Xianping},
  journal={IEEE Transactions on Neural Networks and Learning Systems}, 
  title={Graph-Collaborated Auto-Encoder Hashing for Multiview Binary Clustering}, 
  year={2023},
  volume={},
  number={},
  pages={1-13},
  doi={10.1109/TNNLS.2023.3239033}}

@ARTICLE{9911691,
  author={Wang, Huibing and Jiang, Guangqi and Peng, Jinjia and Deng, Ruoxi and Fu, Xianping},
  journal={IEEE Transactions on Multimedia}, 
  title={Towards Adaptive Consensus Graph: Multi-view Clustering via Graph Collaboration}, 
  year={2022},
  volume={},
  number={},
  pages={1-13},
  doi={10.1109/TMM.2022.3212270}}

@ARTICLE{Person_reid_Ye2022,
  author={Ye, Mang and Shen, Jianbing and Lin, Gaojie and Xiang, Tao and Shao, Ling and Hoi, Steven C. H.},
  journal={IEEE Transactions on Pattern Analysis and Machine Intelligence}, 
  title={Deep Learning for Person Re-Identification: A Survey and Outlook}, 
  year={2022},
  volume={44},
  number={6},
  pages={2872-2893},
  doi={10.1109/TPAMI.2021.3054775}}

@proceedings{proc-iccv-2019,
   city = {Seoul, Korea (South)},
   publisher = {IEEE},
   title = {2019 IEEE/CVF International Conference on Computer Vision (ICCV)},
   editor = {In So Kweon and Nikos Paragios and Ming-Hsuan Yang and Svetlana Lazebnik},
   year = {2019},
}

@proceedings{proc-iccv-2021,
   city = {Seoul, Korea (South)},
   publisher = {IEEE},
   title = {2021 IEEE/CVF International Conference on Computer Vision (ICCV)},
   editor = {Dima Damen and Tal Hassner and Chris Pal and Yoichi Sato},
   year = {2021},
}

@proceedings{proc-icip-2016,
   booktitle={2016 IEEE International Conference on Image Processing (ICIP)},
   year={2016},
   editor = {Lina Karam},
}

@proceedings{proc-icip-2017,
   booktitle={2017 IEEE International Conference on Image Processing (ICIP)},
   year={2017},
   editor = {Xinggang Lin},
}

@proceedings{proc-cvpr-2021,
   booktitle={2021 IEEE/CVF Conference on Computer Vision and Pattern Recognition (CVPR)},
   year={2021},
   editor = {David Forsyth and  Georgia Gkioxari and Tinne Tuytelaars and Ruigang Yang and Jingyi Yu},
}

@proceedings{proc-wacw-2022,
   booktitle={2022 IEEE/CVF Winter Conference on Applications of Computer Vision Workshops (WACVW)},
   year={2022},
   editor = {Ryan Farrell and Catherine Zhao and Saket Anand and Richard Souvenir},
}

@proceedings{proc-cvpr-2019,
   booktitle={2019 IEEE/CVF Conference on Computer Vision and Pattern Recognition (CVPR)},
   year={2019},
   editor = {Abhinav Gupta and Derek Hoiem and Gang Hua and Zhuowen Tu},
}

@proceedings{proc-cvpr-2016,
   booktitle={2016 IEEE/CVF Conference on Computer Vision and Pattern Recognition (CVPR)},
   year={2016},
   editor = {Lourdes Agapito and Tamara Berg and Jana Kosecka and Lihi Zelnik-Manor},
}

@proceedings{proc-cvpr-2012,
   booktitle={2012 IEEE/CVF Conference on Computer Vision and Pattern Recognition (CVPR)},
   year={2012},
   editor = {IEEE},
}

@proceedings{proc-icra-2021,
   booktitle={2021 IEEE International Conference on Robotics and Automation (ICRA)},
   year={2021},
   editor = {IEEE},
}

@proceedings{proc-icme-2018,
   booktitle={2018 IEEE International Conference on Multimedia and Expo (ICME)},
   year={2018},
   editor = {IEEE},
}
\end{document}